%% file: acl2016_1_2.tex
\documentclass[11pt]{article}
\usepackage{acl2016}
\usepackage{times}
\usepackage{latexsym}

\usepackage{amsmath,graphicx} 
\usepackage{algorithm}
\usepackage{algorithmic}
\usepackage{subfigure}

\usepackage{amssymb}
\usepackage{xcolor}
\usepackage{colortbl}

\definecolor{Gray}{gray}{0.8}

\aclfinalcopy 

\title{Deep Reinforcement Learning with a Natural Language Action Space}
\author{%
Ji He$^*$, Jianshu Chen$^\dagger$, Xiaodong He$^\dagger$, Jianfeng Gao$^\dagger$, Lihong Li$^\dagger$\\ {\bf Li Deng$^\dagger$ and Mari Ostendorf$^*$} \\
$^*$Department of Electrical Engineering,
University of Washington,
Seattle, WA 98195, USA \\
\texttt{\{jvking, ostendor\}@uw.edu}
\\
$^\dagger$Microsoft Research,
Redmond, WA 98052, USA \\
\texttt{\{jianshuc, xiaohe, jfgao, lihongli, deng\}@microsoft.com}
}

\date{}

\begin{document}

\maketitle

\begin{abstract}
This paper introduces a novel architecture for reinforcement learning with deep neural networks designed to handle state and action spaces characterized by natural language, as found in text-based games. Termed a deep reinforcement relevance network (DRRN), the architecture represents action and state spaces with separate embedding vectors, which are combined with an interaction function to approximate the Q-function in reinforcement learning. We evaluate the DRRN on two popular text games, showing superior performance over other deep Q-learning architectures. Experiments with paraphrased action descriptions show that the model is extracting meaning rather than simply memorizing strings of text.

\end{abstract}

\section{Introduction}
\label{sec:intro}

\input{intro}

\section{Deep Reinforcement Relevance Network}
\label{sec:dssm-rl}

\input{drrn}

\section{Experimental Results}
\label{sec:experiment}

\input{expts}

\section{Related Work}
\label{sec:related work}

\input{related}

\section{Conclusion}
\label{sec:discussion and future}

In this paper we develop a deep reinforcement relevance network, a novel DNN architecture for handling actions described by natural language in decision-making tasks such as text games. We show that the DRRN converges faster and to a better solution for Q-learning than alternative architectures that do not use separate embeddings for the state and action spaces. Future work includes: (i) adding an attention model to robustly analyze which part of state/actions text correspond to strategic planning, and (ii) applying the proposed methods to more complex text games or other tasks with actions defined through natural language.


\section*{Acknowledgments}

We thank Karthik Narasimhan and Tejas Kulkarni for providing instructions on setting up their parser-based games.

\bibliography{refs}
\bibliographystyle{acl2016}

\end{document}


\section*{\Large Supplementary Material for ``Deep Reinforcement Learning with a Natural Language Action Space''}

\input{appendices}

%% file: intro.tex
This work is concerned with learning strategies for sequential decision-making tasks, where a system takes actions at a particular state with the goal of maximizing a long-term reward. More specifically, we consider tasks where both the states and the actions are characterized by natural language, such as in human-computer dialog systems, tutoring systems, or text-based games. In a text-based game, for example, the player (or system, in this case) is given a text string that describes the current state of the game and several text strings that describe possible actions one could take. After selecting one of the actions, the environment state is updated and revealed in a new textual description. A reward is given either at each transition or in the end. The objective is to understand, at each step, the state text and all the action texts to pick the most relevant action, navigating through the sequence of texts so as to obtain the highest long-term reward.
Here the notion of relevance is based on the joint state/action impact on the reward: an action text string is said to be ``more relevant'' (to a state text string) than the other action texts if taking that action would lead to a higher long-term reward. Because a player's action changes the environment, reinforcement learning \cite{sutton1998reinforcement} is appropriate for modeling long-term dependency in text games.

 There is a large body of work on reinforcement learning. Of most interest here are approaches leveraging neural networks because of their success in handling a large state space.
Early work --- TD-gammon --- used a neural network to approximate the state value function \cite{tesauro1995temporal}. Recently, inspired by advances in deep learning \cite{lecun2015deep,hinton2012deep,krizhevsky2012imagenet,dahl2012context}, significant progress has been made by combining deep learning with reinforcement learning.  Building on the approach of Q-learning \cite{watkins1992q}, the ``Deep Q-Network'' (DQN) was developed and applied to Atari games \cite{2013arXiv1312.5602M,mnih2015human} and shown to achieve human level performance by applying convolutional neural networks to the raw image pixels.  Narasimhan et al.~\shortcite{narasimhan-kulkarni-barzilay:2015:EMNLP} applied a Long Short-Term Memory network to characterize the state space in a DQN framework for learning control policies for parser-based text games. More recently, Nogueira and Cho~\shortcite{nogueira2016webnav} have also proposed a goal-driven web navigation task for language based sequential decision making study. Another stream of work focuses on continuous control with deep reinforcement learning \cite{lillicrap2015continuous}, where an actor-critic algorithm operates over a known continuous action space.

Inspired by these successes and recent work using neural networks to learn phrase- or sentence-level embeddings 
\cite{collobert2008unified,huang2013learning,le2014distributed,sutskever2014sequence,kiros2015skip}, 
we propose a novel deep architecture for text understanding, which we call a deep
reinforcement relevance network (DRRN). The DRRN uses separate deep neural networks to map state and action text strings into  embedding vectors, from which ``relevance'' is measured numerically by a general interaction function, such as their inner product. The output of this interaction function defines the value of the Q-function for the current state-action pair, which characterizes the optimal long-term reward for pairing these two text strings. The Q-function approximation is learned in an end-to-end manner by Q-learning. 

The DRRN differs from prior work in that earlier studies mostly considered action spaces that are bounded and known. For actions described by natural language text strings, the action space is inherently discrete and potentially unbounded due to the exponential complexity of language with respect to sentence length. A distinguishing aspect of the DRRN architecture --- compared to simple DQN extensions --- is that two different types of meaning representations are learned, reflecting the tendency for state texts to describe scenes and action texts to describe potential actions from the user.  We show that the DRRN learns a continuous space representation of actions that successfully generalize to paraphrased descriptions of actions unseen in training.


%% file: drrn.tex
\subsection{Text Games and Q-learning}


We consider the sequential decision making problem for text understanding. At each time step $t$, the agent will receive a string of text that describes the state $s_t$ (i.e., ``state-text'') and several strings of text that describe all the potential actions $a_t$ (i.e., ``action-text''). The agent attempts to understand the texts from both the state side and the action side, measuring their relevance to the current context $s_t$ for the purpose of maximizing the long-term reward, and then picking the best action. Then, the environment state is updated $s_{t+1}=s'$ according to the probability $p(s'|s,a)$, and the agent receives a reward $r_t$ for that particular transition. The \emph{policy} of the agent is defined to be the probability $\pi(a_t|s_t)$ of taking action $a_t$ at state $s_t$. Define the Q-function $Q^{\pi}(s,a)$ as the expected return starting from $s$, taking the action $a$, and thereafter following policy $\pi(a|s)$ to be:
$$		Q^{\pi}(s,a) = \mathbb{E}\left\{\sum_{k=0}^{+\infty} \gamma^k r_{t+k} \bigg| s_t = s, a_t = a\right\}
$$
where $\gamma$ denotes a discount factor.
The optimal policy and Q-function can be found by using the Q-learning algorithm \cite{watkins1992q}:
	\begin{align}
		\label{Equ:DRRN:QLearning_Recursion}
		Q(s_t, a_t) 
			&\leftarrow
				Q(s_t, a_t) 
				+ \\ \nonumber
			&	\eta_t \cdot 
				\big(
					r_t + \gamma \cdot \max_{a}{Q(s_{t+1}, a)}-Q(s_t, a_t)
				\big) 
	\end{align}
where $\eta_t$ is the learning rate of the algorithm. 
In this paper, we use a softmax selection strategy as the exploration policy during the learning stage, which chooses the action $a_t$ at state $s_t$ according to the following probability:
	\begin{align}
		\pi(a_t = a_t^i | s_t)	
				&=
					\frac{\exp(\alpha \cdot Q(s_t, a_t^i))}
						{\sum_{j=1}^{|\mathcal{A}_t|} \exp( \alpha \cdot  Q(s_t, a_t^j))},
		\label{Equ:DRRN:SoftmaxSelection}
	\end{align}
where $\mathcal{A}_t$ is the set of feasible actions at state $s_t$, $a_t^i$ is the $i$-th feasible action in $\mathcal{A}_t$, $|\cdot|$ denotes the cardinality of the set, and $\alpha$ is the scaling factor in the softmax operation.
$\alpha$ is kept constant throughout the learning period. All methods are initialized with small random weights, so initial Q-value differences will be small, thus making the Q-learning algorithm more explorative initially. As Q-values better approximate the true values, a reasonable $\alpha$ will make action selection put high probability on the optimal action (exploitation), but still maintain a small exploration probability.

\subsection{Natural language action space}

Let $\mathcal{S}$ denote the state space, and let $\mathcal{A}$ denote the entire action space that includes all the unique actions over time. A vanilla Q-learning recursion \eqref{Equ:DRRN:QLearning_Recursion} needs to maintain a table of size $| \mathcal{S} | \times | \mathcal{A} |$, which is problematic for a large state/action space. Prior work using a DNN in Q-function approximation has shown high capacity and scalability for handling a large state space, but most studies  have used a network that generates $|\mathcal{A}|$ outputs, each of which represents the value of $Q(s,a)$ for a particular action $a$. 
It is not practical to have a DQN architecture of a size that is explicitly dependence on the large number of natural language actions.
 Further, in many text games, the feasible action set $\mathcal{A}_t$ at each time $t$ is an unknown subset of the unbounded action space $\mathcal{A}$ that varies over time.  

\begin{figure*}[t]
  \centerline{
	\subfigure[Max-action DQN]{
  	\includegraphics[width=5.6cm]{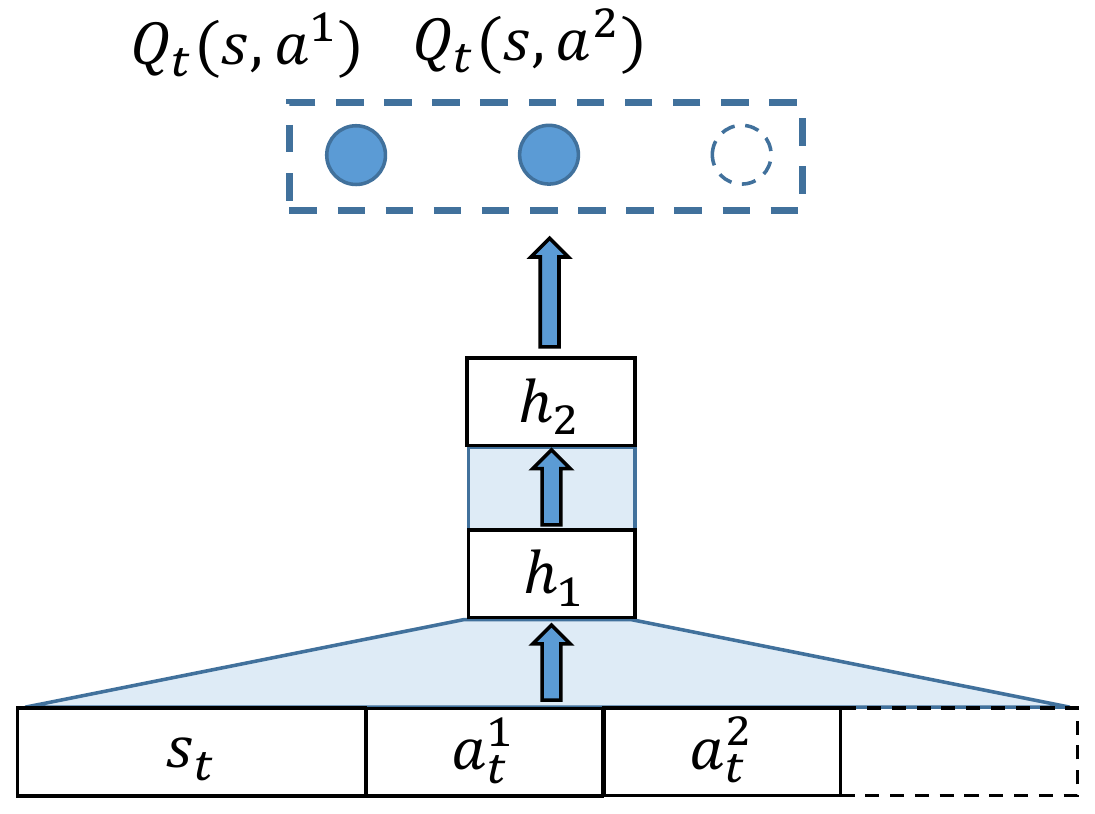}
  	\label{fig:architectures_dqn}
  	}
  	\hfill
	\subfigure[Per-action DQN]{
	\includegraphics[width=3.2cm]{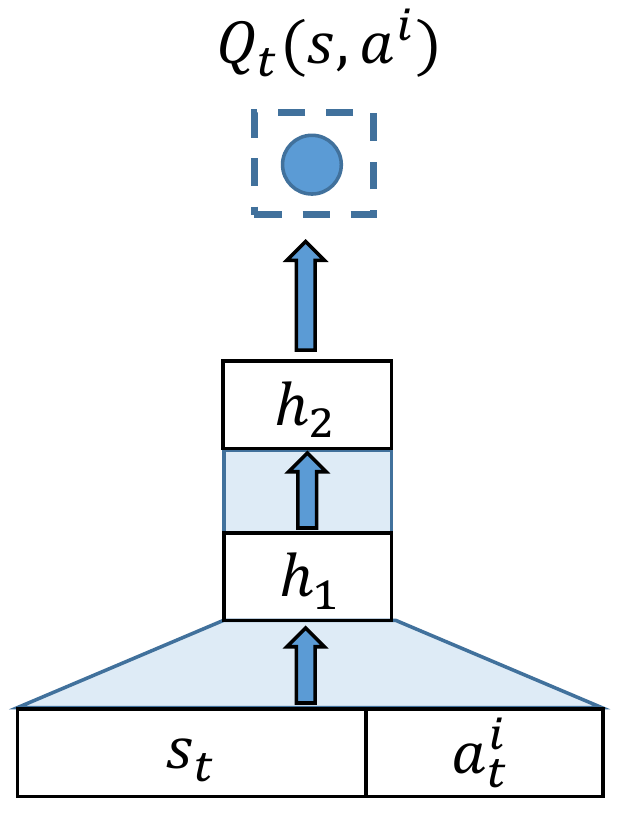}
	\label{fig:architectures_nn-rl}
  	}
  	\hfill
	\subfigure[DRRN]{
	\includegraphics[width=6.1cm]{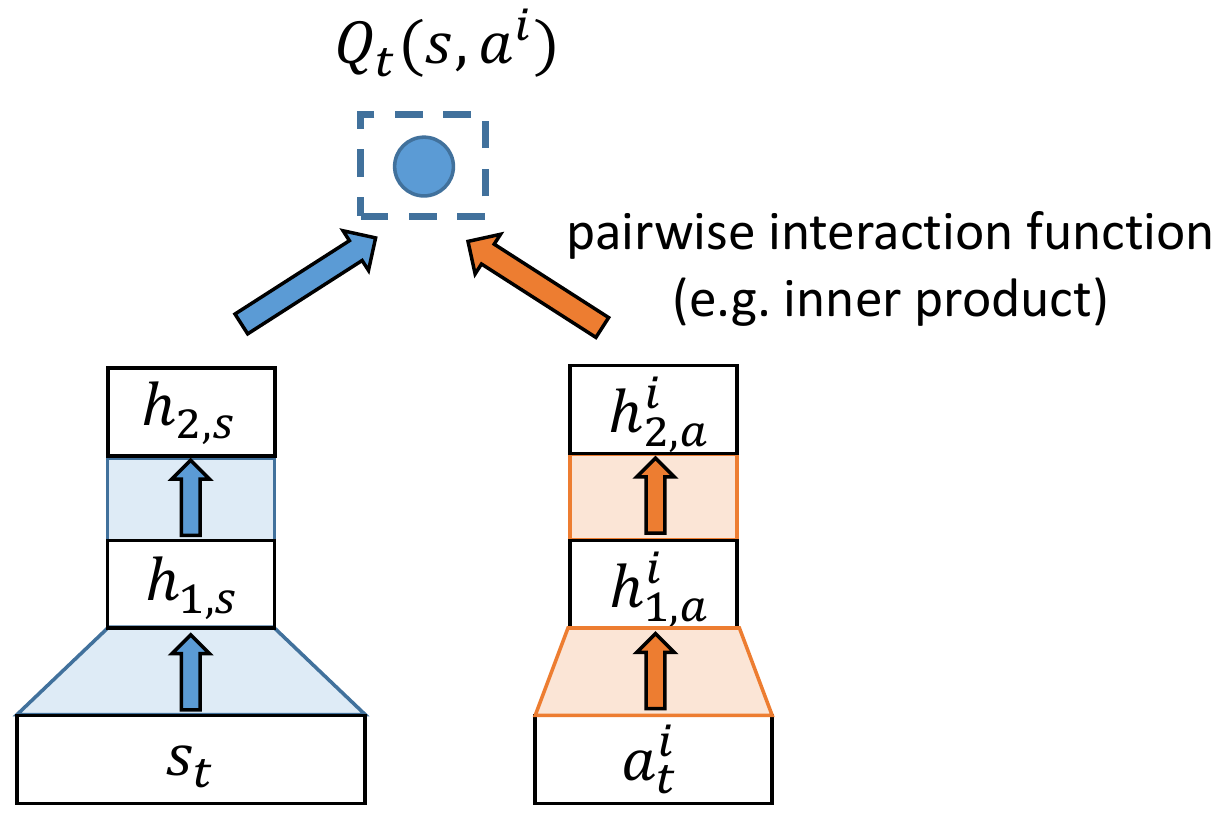}
  	\label{fig:architectures_drrn}
  	}
	}
\caption{Different deep Q-learning architectures: Max-action DQN and Per-action DQN both treat input text as concantenated vectors and compute output Q-values with a single NN. DRRN models text embeddings from state/action sides separately, and use an interaction function to compute Q-values.}
\label{fig:architectures}
\end{figure*}

For the case where the maximum number of possible actions at any point in time ($\max_{t} |\mathcal{A}_t|$) is known, the DQN can be modified to simply use that number of outputs (``Max-action DQN''), as
illustrated in Figure \ref{fig:architectures_dqn}, where the state and action vectors are concatenated (i.e., as an extended state vector) as its input. The network computes the Q-function values for the actions in the current feasible set as its outputs.  For a complex game, $\max_{t} |\mathcal{A}_t|$ may be difficult to obtain, because $\mathcal{A}_t$ is usually unknown beforehand. Nevertheless, we will use this modified DQN as a baseline.

An alternative approach is to use a function approximation using a neural network that takes a state-action pair as input, and outputs a single Q-value for each possible action (``Per-action DQN'' in Figure \ref{fig:architectures_nn-rl}). This architecture easily handles a varying number of actions and represents a second baseline.

We propose an alternative architecture for handling a natural language action space in sequential text understanding: the deep reinforcement relevance network (DRRN). As shown in Figure \ref{fig:architectures_drrn}, the DRRN consists of a pair of DNNs, one for the state text embedding and the other for action text embeddings, which are combined using a pairwise interaction function. The texts used to describe states and actions could be very different in nature, e.g., a state text could be long, containing sentences with complex linguistic structure, whereas an action text could be very concise or just a verb phrase. Therefore, it is desirable to use two networks with different structures to handle state/action texts, respectively. As we will see in the experimental sections, by using two separate deep neural networks for state and action sides, we obtain much better results.

\subsection{DRRN architecture: Forward activation}
\label{subsec:drrn-forward-activation}

Given any state/action text pair $(s_t, a_t^i)$, the DRRN estimates the Q-function $Q(s_t, a_t^i)$ in two steps. First, map both $s_t$ and $a_t^i$ to their embedding vectors using the corresponding DNNs, respectively. Second, approximate $Q(s_t, a_t^i)$ using an interaction function such as the inner product of the embedding vectors. Then, given a particular state $s_t$, we can select the optimal action $a_t$ among the set of actions via $a_t = \arg\,\max_{a_t^i} Q(s_t, a_t^i)$. 

More formally, let $h_{l,s}$ and $h_{l,a}$ denote the $l$-th hidden layer for state and action side neural networks, respectively. For the state side, $W_{l,s}$ and $b_{l,s}$ denote the linear transformation weight matrix and bias vector between the $(l-1)$-th and $l$-th hidden layers. $W_{l,a}$ and $b_{l,a}$ denote the equivalent parameters for the action side. In this study, the DRRN has $L$ hidden layers on each side.
	\begin{align}
		h_{1,s}			&=	f(W_{1,s} s_t + b_{1,s}) \\
		h_{1,a}^i			&=	f(W_{1,a} a_t^i + b_{1,a}) \\
		h_{l,s}			&=	f(W_{l-1,s} h_{l-1,s} + b_{l-1,s})  \\
		h_{l,a}^i			&=	f(W_{l-1,a} h_{l-1,a}^i + b_{l-1,a})
	\end{align}
where $f(\cdot)$ is the nonlinear activation function at the hidden layers, which, for example, could be chosen as $\tanh{(x)}$, and $i=1, 2, 3, ..., |\mathcal{A}_t|$ is the action index.
A general interaction function $g(\cdot)$ is used to approximate the Q-function values, $Q(s, a)$, in the following parametric form:
	\begin{align}
		Q(s, a^i; \Theta)=
					g\left(h_{L,s},\ h_{L,a}^i\right)
		\label{Equ:DRRN:Forward:InnerProduct}
	\end{align}
where $\Theta$ denotes all the model parameters. The interaction function could be an inner product, a bilinear operation, or a nonlinear function such as a deep neural network. In our experiments, the inner product and bilinear operation gave similar results. For simplicity, we present our experiments mostly using the inner product interaction function.

The success of the DRRN in handling a natural language action space $\mathcal{A}$ lies in the fact that the state-text and the action-texts are mapped into separate finite-dimensional embedding spaces. The end-to-end learning process (discussed next) makes the embedding vectors  in the two spaces more aligned for ``good'' (or relevant) action texts compared to ``bad'' (or irrelevant) choices, resulting in a higher interaction function output (Q-function value).

\subsection{Learning the DRRN: Back propagation}
\label{Sec:DRRN:Learning}

\begin{figure*}[t]
  \centering
  \includegraphics[width=14cm]{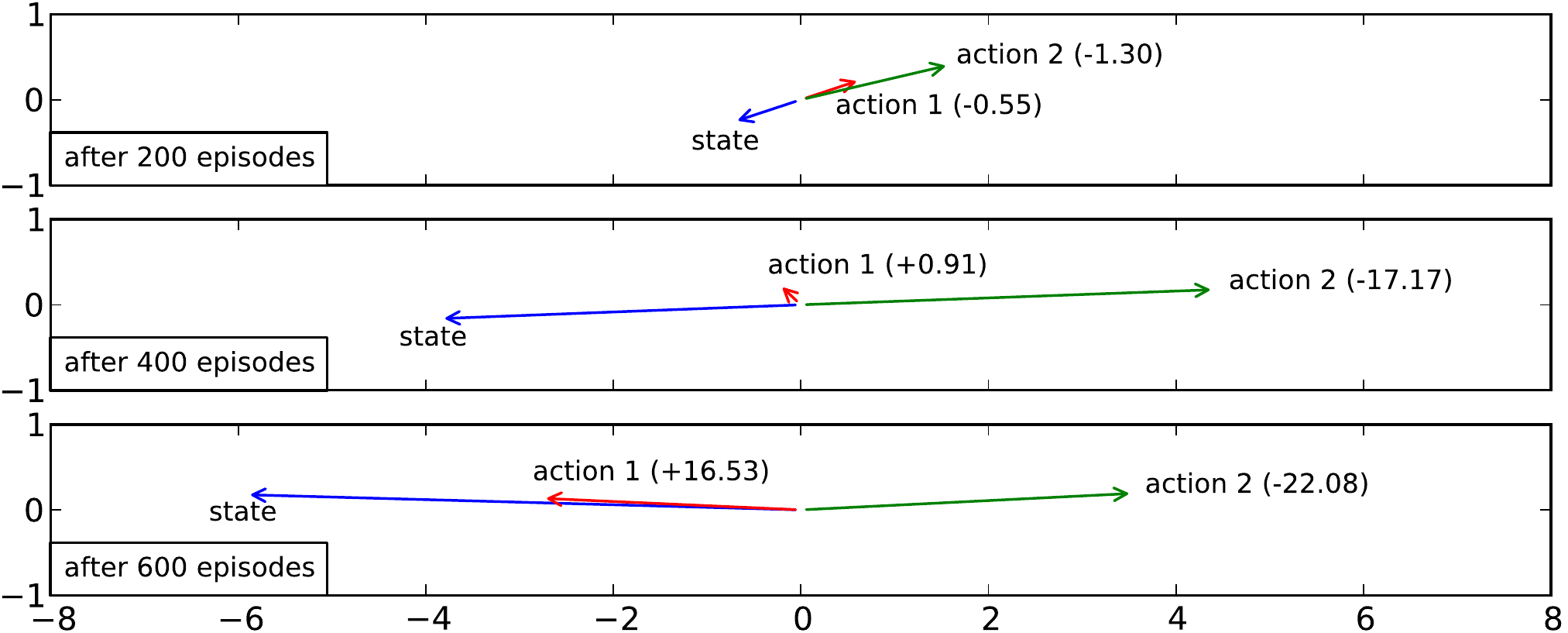}
\caption{PCA projections of text embedding vectors for state and associated action vectors after 200, 400 and 600 training episodes. The state is ``As you move forward, the people surrounding you suddenly look up with terror in their faces, and flee the street.'' 
Action 1 (good choice) is ``Look up'', and action 2 (poor choice) is ``Ignore the alarm of others and continue moving forward.'' 
}
\label{Fig:visual-embed-trace}
\end{figure*}

To learn the DRRN, we use the ``experience-replay'' strategy \cite{lin1993reinforcement}, which uses a fixed exploration policy to interact with the environment to obtain a sample trajectory. Then, we randomly sample a transition tuple $(s_k, a_k, r_k, s_{k+1})$, compute the temporal difference error for sample $k$:
$$d_k=r_k+\gamma \max_{a}{Q(s_{k+1}, a; \Theta_{k-1})}-Q(s_k, a_k; \Theta_{k-1}),$$
and update the model according to the recursions: 
	\begin{align}
		W_{v,k}		&=
						W_{v,k-1} + \eta_k d_k \cdot
						\frac{\partial Q(s_k, a_k; \Theta_{k-1})}{\partial W_v}	\\
		b_{v,k}		&=
						b_{v,k-1} + \eta_k d_k \cdot
						\frac{\partial Q(s_k, a_k; \Theta_{k-1})}{\partial b_v}
		\label{Equ:DRRN:Backward}
	\end{align}
for $v\in\{s,a\}$. Expressions for $\frac{\partial Q}{\partial W_v}$, $\frac{\partial Q}{\partial b_v}$ and other algorithm details are given in supplementary materials. Random sampling essentially scrambles the trajectory from experience-replay into a ``bag-of-transitions'', which has been shown to avoid oscillations or divergence and achieve faster convergence in Q-learning \cite{mnih2015human}.  Since the models on the action side share the same parameters, models associated with all actions are effectively updated even though the back propagation is only over one action.
We apply back propagation to learn how to pair the text strings from the reward signals in an end-to-end manner. The representation vectors for the state-text and the action-text are \emph{automatically learned} to be aligned with each other in the text embedding space from the reward signals. A summary of the full learning algorithm is given in Algorithm \ref{Alg:DRRN}.

\begin{algorithm*}[t!]
\begin{algorithmic}[1]
\STATE Initialize replay memory $\mathcal{D}$ to capacity $N$.
\STATE Initialize DRRN with small random weights.
\STATE Initialize game simulator and load dictionary.
\FOR{$episode = 1, \ldots, M$}
	\STATE Restart game simulator.
	\STATE Read raw state text and a list of action text from the simulator, and convert them to representation $s_1$ and $a_1^1, a_1^2, \ldots, a_1^{|\mathcal{A}_1|}$.
	\FOR{$t = 1, \ldots, T$}
		\STATE Compute $Q(s_t, a_t^i; \Theta)$ for the list of actions using DRRN forward activation (Section \ref{subsec:drrn-forward-activation}).
		\STATE Select an action $a_t$ based on probability distribution $\pi(a_t = a_t^i | s_t)$ (Equation \ref{Equ:DRRN:SoftmaxSelection})
		\STATE Execute action $a_t$ in simulator
		\STATE Observe reward $r_t$. Read the next state text and the next list of action texts, and convert them to representation $s_{t+1}$ and $a_{t+1}^1, a_{t+1}^2, \ldots, a_{t+1}^{|\mathcal{A}_{t+1}|}$.
		\STATE Store transition $(s_t, a_t, r_t, s_{t+1}, A_{t+1})$ in $\mathcal{D}$.
		\STATE Sample random mini batch of transitions $(s_k, a_k, r_k, s_{k+1}, A_{k+1})$ from $\mathcal{D}$.
		\STATE Set $y_k=\begin{cases}
					r_k & \text{if } s_{k+1} \text{ is terminal}\\
					r_k+\gamma\max_{a'\in A_{k+1}}Q(s_{k+1}, a'; \Theta)) & \text{otherwise}
				\end{cases}$
		\STATE Perform a gradient descent step on $(y_k-Q(s_k, a_k; \Theta))^2$ with respect to the network parameters $\Theta$ (Section \ref{Sec:DRRN:Learning}). Back-propagation is performed only for $a_k$ even though there are $|\mathcal{A}_{k}|$ actions at time $k$.
\ENDFOR
\ENDFOR
\end{algorithmic}
\caption{Learning algorithm for DRRN}
\label{Alg:DRRN}
\end{algorithm*}

Figure \ref{Fig:visual-embed-trace} illustrates learning with an inner product interaction function. We used Principal Component Analysis (PCA) to project the 100-dimension last hidden layer representation (before the inner product) to a 2-D plane. The vector embeddings start with small values, and after 600 episodes of experience-replay training, the embeddings are very close to the converged embedding (4000 episodes). The embedding vector of the optimal action (Action 1) converges to a positive inner product with the state embedding vector, while Action 2 converges to a negative inner product.

%% file: expts.tex
\subsection{Text games}
\label{sec:experiment:TextGame}

\begin{figure*}[t]
  \centerline{
	\subfigure[Parser-based]{
	\includegraphics[width=0.32\textwidth]{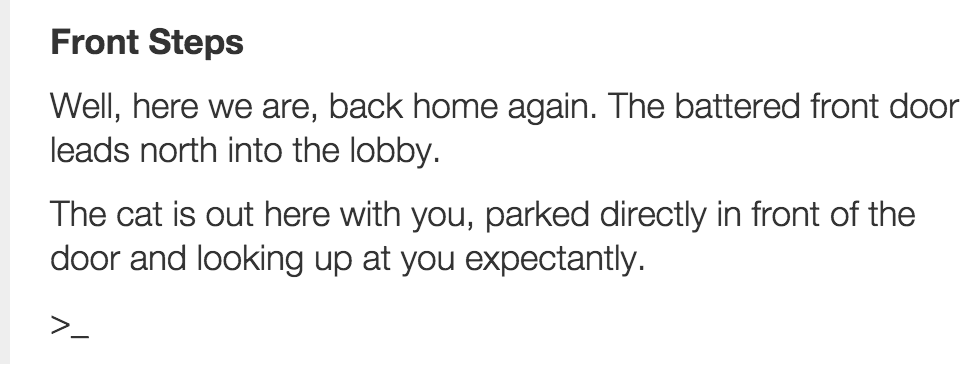}
	\label{fig:parser-based}
  	}
	\subfigure[Choiced-based]{
  	\includegraphics[width=0.32\textwidth]{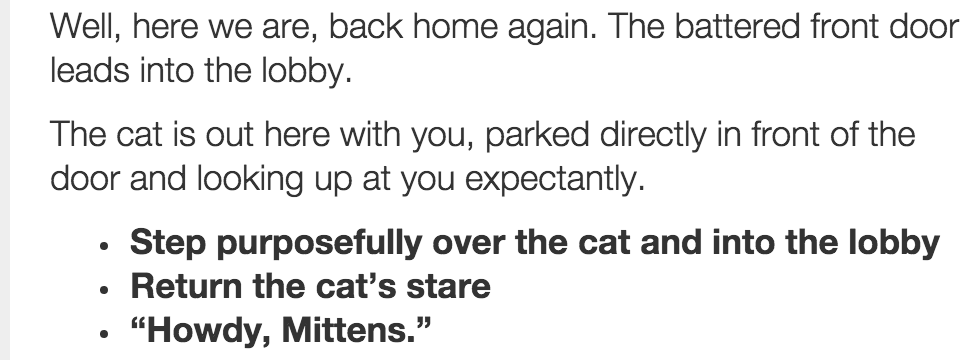}
  	\label{fig:choice-based}
  	}
	\hfil
	\subfigure[Hypertext-based]{
	\includegraphics[width=0.32\textwidth]{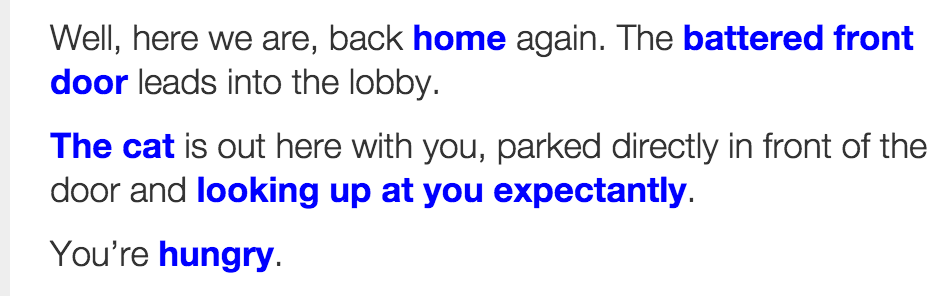}
 	\label{fig:hypertext-based}
  	}
	}
\caption{Different types of text games}
\end{figure*}

Text games, although simple compared to video games, still enjoy high popularity in online communities, with annual competitions held online since 1995. Text games communicate to players in the form of a text display, which players have to understand and respond to by typing or clicking text \cite{adams2014fundamentals}. There are three types of text games: 
parser-based (Figure \ref{fig:parser-based}), choice-based (Figure \ref{fig:choice-based}), and hypertext-based (Figure \ref{fig:hypertext-based}). 
Parser-based games accept typed-in commands from the player, usually in the form of verb phrases, such as ``eat apple'', ``get key'', or ``go east''. 
They involve the least complex action language.
Choice-based and hypertext-based games present actions after or embedded within the state text. The player chooses an action, and the story continues based on the action taken at this particular state. With the development of web browsing and richer HTML display, choice-based and hypertext-based text games have become more popular, increasing in percentage from 8\% in 2010 to 62\% in 2014.\footnote{Statistics obtained from \tt{http://www.ifarchive.
org}}

For parser-based text games, Narasimhan et al.~\shortcite{narasimhan-kulkarni-barzilay:2015:EMNLP} have defined a fixed set of 222 actions, which is the total number of possible phrases the parser accepts. Thus the parser-based text game is reduced to a problem that is
well suited to a fixed-action-set DQN.
However, for choice-based and hypertext-based text games, the size of the action space could be exponential with the length of the action sentences, which is handled here by using a continuous representation of the action space.


\begin{table}[t]
\begin{center}
\footnotesize
\begin{tabular}{| p{3.0cm} | p{1.7cm} | p{1.5cm} |} \hline
\rowcolor{Gray} Game & Saving John & Machine of Death \\ \hline
Text game type & Choice & Choice \& Hypertext \\ \hline
Vocab size & 1762 & 2258  \\ \hline
Action vocab size & 171 & 419 \\ \hline
Avg. words/description & 76.67 & 67.80 \\ \hline
State transitions & Deterministic & Stochastic \\ \hline
\# of states (underlying) & $\ge 70$ & $\ge 200$ \\ \hline
\end{tabular}
\caption{Statistics for the games ``Saving John'' and and ``Machine of Death''.}
\label{table:basic-stats}
\end{center}
\end{table}

In this study, we evaluate the DRRN with two games: a deterministic text game task called ``Saving John'' and a larger-scale stochastic text game called ``Machine of Death'' from a public archive.\footnote{Simulators are available at \tt{https://github.com/
jvking/text-games}} The basic text statistics of these tasks are shown in Table \ref{table:basic-stats}. The maximum value of feasible actions (i.e., $\max_t |\mathcal{A}_t|$) is four in ``Saving John'', and nine in ``Machine of Death''. We manually annotate final rewards for all distinct endings in both games (as shown in supplementary materials). The magnitude of reward scores are given to describe sentiment polarity of good/bad endings. On the other hand, each non-terminating step we assign with a small negative reward, to encourage the learner to finish the game as soon as possible. For the text game ``Machine of Death'', we restrict an episode to be no longer than 500 steps. 

In ``Saving John'' all actions are choice-based, for which the mapping from text strings to $a_t$ are clear. In ``Machine of Death'', when actions are hypertext, the actions are substrings of the state. In this case $s_t$ is associated with the full state description, and $a_t$ are given by the substrings without any surrounding context. For text input, we use raw bag-of-words as features, with different vocabularies for the state side and action side. 


\subsection{Experiment setup}
We apply DRRNs with both 1 and 2 hidden layer structures. In most experiments, we use dot-product as the interaction function and set the hidden dimension to be the same for each hidden layer. 
We use DRRNs with 20, 50 and 100-dimension hidden layer(s) and build learning curves during experience-replay training. The learning rate is constant: $\eta_t=0.001$. In testing, as in training, we apply softmax selection. We record average final rewards as performance of the model. 


The DRRN is compared to multiple baselines: a linear model, two max-action DQNs (MA DQN) ($L=1$ or 2 hidden layers), and two per-action DQNs (PA DQN) (again, $L=1,2$). All baselines use the same Q-learning framework with different function approximators to predict $Q(s_t, a_t)$ given the current state and actions. For the linear and MA DQN baselines, the input is the text-based state and action descriptions, each as a bag of words, with the number of outputs equal to the maximum number of actions. When there are fewer actions than the maximum, the highest scoring available action is used. The PA DQN baseline takes each pair of state-action texts as input, and generates a corresponding Q-value.

We use softmax selection, which is widely applied in practice, to trade-off exploration vs. exploitation. Specifically, for each experience-replay, we first generate 200 episodes of data (about 3K tuples in ``Saving John'' and 16K tuples in ``Machine of Death'') using the softmax selection rule in \eqref{Equ:DRRN:SoftmaxSelection}, where we set $\alpha=0.2$ for the first game and $\alpha=1.0$ for the second game. The $\alpha$ is picked according to an estimation of range of the optimal Q-values. We then shuffle the generated data tuples $(s_t, a_t, r_t, s_{t+1})$ update the model as described in Section \ref{Sec:DRRN:Learning}. The model is trained with multiple epochs for all configurations, and is evaluated after each experience-replay. The discount factor $\gamma$ is set to 0.9. For DRRN and all baselines, network weights are initialized with small random values. To prevent algorithms from ``remembering'' state-action ordering and make choices based on action wording, each time the algorithm/player reads text from the simulator, we randomly shuffle the list of actions.\footnote{When in a specific state, the simulator presents the possible set of actions in random order, i.e. they may appear in a different order the next time a player is in this same state.}
This will encourage the algorithms to make decisions based on the understanding of the texts that describe the states and actions.


\subsection{Performance}

\begin{figure*}[t]
\centerline{
	\subfigure[Game 1: ``Saving John'']{
		\includegraphics[width=0.48\textwidth]{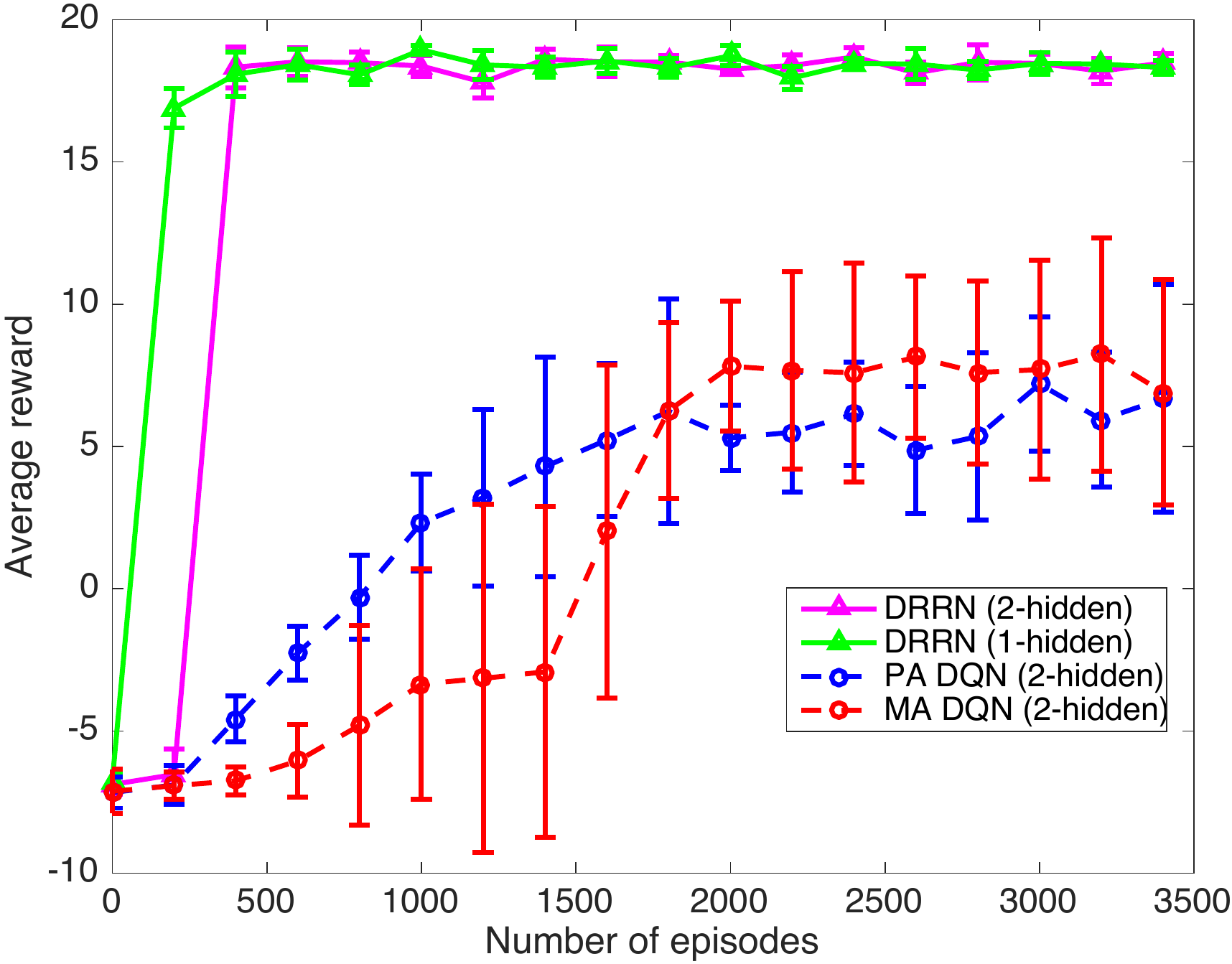}
		 \label{fig:learning_curve1}
		}
	\hfil
	\subfigure[Game 2: ``Machine of Death'']
	{
		\includegraphics[width=0.48\textwidth]{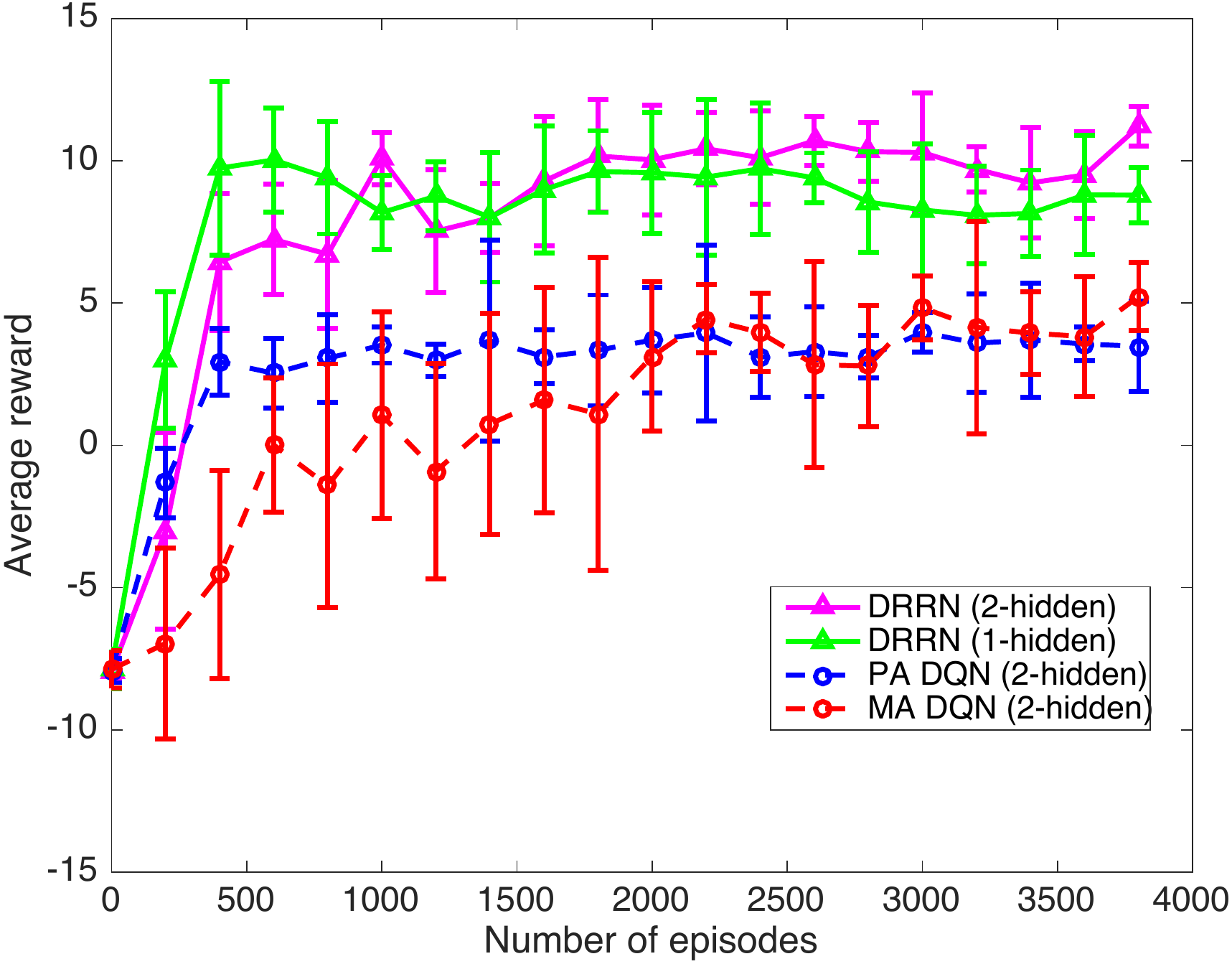}
		\label{fig:learning_curve2}
	}
 } 
  \caption{Learning curves of the two text games.}
  \label{Fig:LearningCurve}
\end{figure*}

\begin{table}[t]
\begin{center}
\footnotesize
\begin{tabular}{| p{2.4cm} | c | c | c |} \hline
\rowcolor{Gray} Eval metric & \multicolumn{3}{c|}{Average reward} \\ \hline
hidden dimension & 20 & 50 & 100 \\ \hline
Linear & \multicolumn{3}{c|}{4.4 (0.4)} \\ \hline
PA DQN ($L=1$) & 2.0 (1.5) & 4.0 (1.4) & 4.4 (2.0) \\ \hline
PA DQN ($L=2$) & 1.5 (3.0) & 4.5 (2.5) & 7.9 (3.0) \\ \hline
MA DQN ($L=1$) & 2.9 (3.1) & 4.0 (4.2) & 5.9 (2.5)  \\ \hline
MA DQN ($L=2$) & 4.9 (3.2) & 9.0 (3.2) & 7.1 (3.1)  \\ \hline
DRRN ($L=1$) & 17.1 (0.6) & 18.3 (0.2) & 18.2 (0.2)  \\ \hline
DRRN ($L=2$) & 18.4 (0.1) & 18.5 (0.3) & {\bf 18.7} (0.4)  \\ \hline
\end{tabular}
\caption{The final average rewards and standard deviations on ``Saving John''.}
\label{table:final-perf}
\end{center}
\end{table}

\begin{table}[t]
\begin{center}
\footnotesize
\begin{tabular}{| p{2.4cm} | >{\centering}p{1.25cm} | c | c |} \hline
\rowcolor{Gray} Eval metric & \multicolumn{3}{c|}{Average reward} \\ \hline
hidden dimension & 20 & 50 & 100 \\ \hline
Linear & \multicolumn{3}{c|}{3.3 (1.0)} \\ \hline
PA DQN ($L=1$) & 0.9 (2.4) & 2.3 (0.9) & 3.1 (1.3) \\ \hline
PA DQN ($L=2$) & 1.3 (1.2) & 2.3 (1.6) & 3.4 (1.7) \\ \hline
MA DQN ($L=1$) & 2.0 (1.2) & 3.7 (1.6) & 4.8 (2.9)  \\ \hline
MA DQN ($L=2$) & 2.8 (0.9) & 4.3 (0.9) & 5.2 (1.2) \\ \hline
DRRN ($L=1$) & 7.2 (1.5) & 8.4 (1.3) & 8.7 (0.9)  \\ \hline
DRRN ($L=2$) & 9.2 (2.1) & 10.7 (2.7) & {\bf 11.2} (0.6)  \\ \hline
\end{tabular}
\caption{The final average rewards and standard deviations on ``Machine of Death''.}
\label{table:final-perf-game2}
\end{center}
\end{table}


In Figure  \ref{Fig:LearningCurve}, we show the learning curves of different models, where the dimension of the hidden layers in the DQNs and DRRN are all set to 100. The error bars are obtained by running 5 independent experiments. The proposed methods and baselines all start at about the same performance (roughly -7 average rewards for Game 1, and roughly -8 average rewards for Game 2), which is the random guess policy. After around 4000 episodes of experience-replay training, all methods converge. The DRRN converges much faster than the other three baselines and achieves a higher average reward. We hypothesize this is because the DRRN architecture is better at capturing relevance between state text and action text.  The faster convergence for ``Saving John'' may be due to the smaller observation space and/or the deterministic nature of its state transitions (in contrast to the stochastic transitions in the other game).


The final performance (at convergence) for both baselines and proposed methods are shown in Tables \ref{table:final-perf} and \ref{table:final-perf-game2}. We test for different model sizes with 20, 50, and 100 dimensions in the hidden layers. The DRRN performs consistently better than all baselines, and often with a lower variance. For Game 2, due to the complexity of the underlying state transition function, we cannot compute the exact optimal policy score.
To provide more insight into the performance, we averaged scores of 8 human players for initial trials (novice) and after gaining experience, yielding scores of $-5.5$ and 16.0, respectively. The experienced players do outperform our algorithm. The converged performance is higher with two hidden layers for all models. However, deep models also converge more slowly than their 1 hidden layer versions, as shown for the DRRN in Figure \ref{Fig:LearningCurve}.

Besides an inner-product, we also experimented with more complex interaction functions: a) a bilinear operation with different action side dimensions; and b) a non-linear deep neural network using the concatenated state and action space embeddings as input and trained in an end-to-end fashion to predict Q values. For different configurations, we fix the state side embedding to be 100 dimensions and vary the action side embedding dimensions. The bilinear operation gave similar results, but the concatenation input to a DNN degraded performance. 
Similar behaviors have been observed on a different task \cite{luong-pham-manning:2015:EMNLP}.



\subsection{Actions with paraphrased descriptions}
To investigate how our models handle actions with ``unseen'' natural language descriptions, we had two people paraphrase all actions in the game ``Machine of Death'' (used in testing phase), except a few single-word actions whose synonyms are out-of-vocabulary (OOV). The word-level OOV rate of paraphrased actions is 18.6\%, and standard 4-gram BLEU score between the paraphrased and original actions is 0.325. The resulting 153 paraphrased action descriptions are associated with 532 unique state-action pairs.

\begin{figure}[t]
	\includegraphics[width=1.0\linewidth]{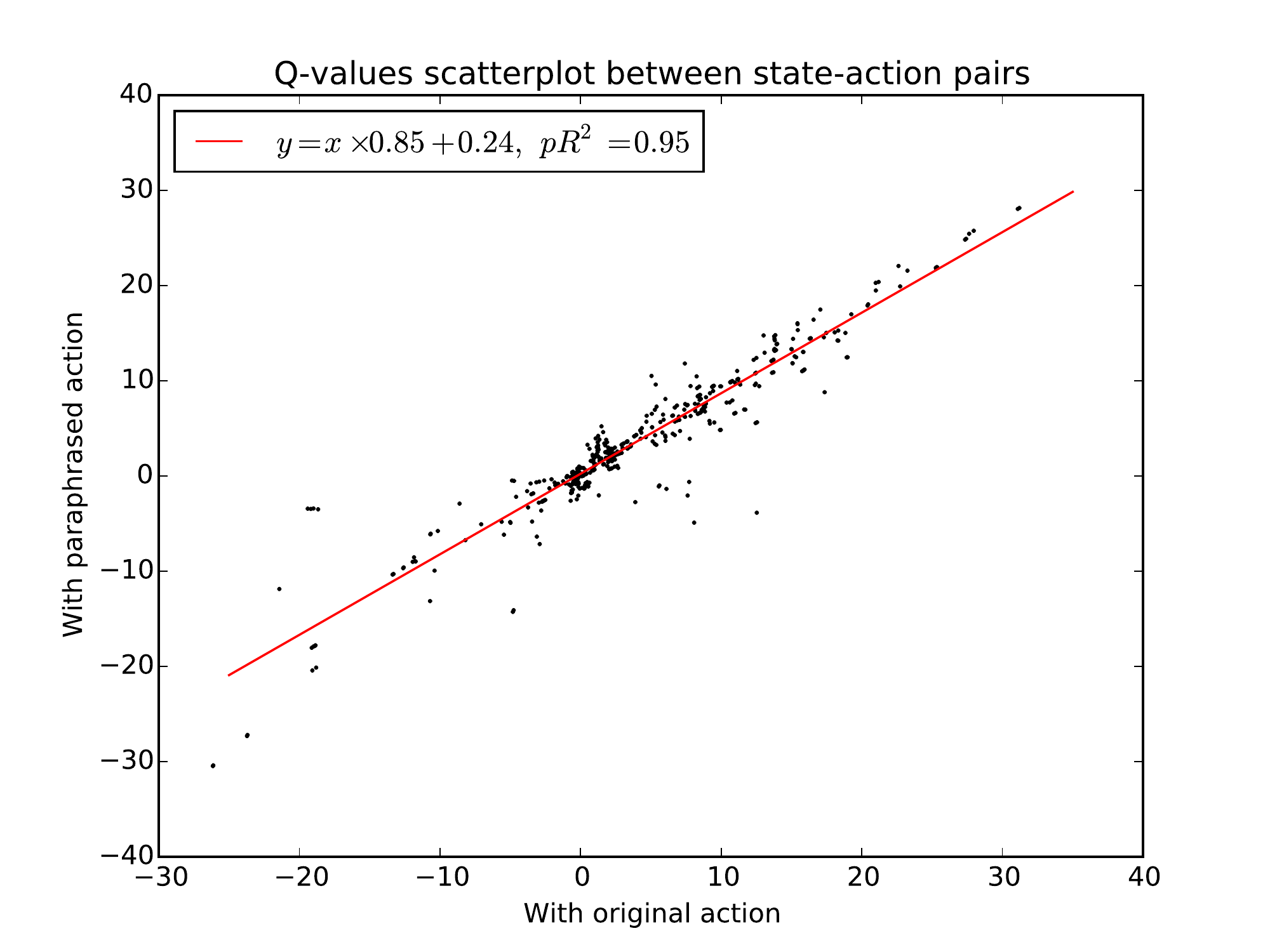}
	\caption{Scatterplot and strong correlation between Q-values of paraphrased actions versus original actions}
	\label{Fig:paraphrase_correlation}
\end{figure}

We apply a well-trained 2-layer DRRN model (with hidden dimension 100), and predict Q-values for each state-action pair with fixed model parameters. Figure \ref{Fig:paraphrase_correlation} shows the correlation between Q-values associated with paraphrased actions versus original actions. The predictive R-squared is 0.95, showing a strong positive correlation. We also run Q-value correlation for the NN interaction and $pR^2=0.90$. For baseline MA-DQN and PA-DQN, their corresponding $pR^2$ is 0.84 and 0.97, indicating they also have some generalization ability. This is confirmed in the paraphrasing-based experiments too, where the test reward on the paraphrased setup is close to the original setup. 
This supports the claim that deep learning is useful in general for this language understanding task, and our findings show that a decoupled architecture most effectively leverages that approach.

In Table \ref{table:q_value_examples} we provide examples with predicted Q-values of original descriptions and paraphrased descriptions. We also 
include alternative action descriptions with in-vocabulary words that 
will lead to positive / negative / irrelevant game development at that particular state. Table \ref{table:q_value_examples} shows actions that are more likely to result in good endings are predicted with high Q-values. This indicates that the DRRN has some generalization ability and gains a useful level of language understanding in the game scenario.

\begin{table*}[t]
\begin{center}
\footnotesize
\begin{tabular}{| p{4.9cm} | p{10.1cm} |} \hline
\rowcolor{Gray}  & Text (with predicted Q-values) \\ \hline
State & As you move forward, the people surrounding you suddenly look up with terror in their faces, and flee the street. \\ \hline
Actions in the original game & Ignore the alarm of others and continue moving forward. (-21.5) Look up. (16.6) \\ \hline
Paraphrased actions (not original) & Disregard the caution of others and keep pushing ahead. (-11.9) Turn up and look. (17.5) \\ \hline
Positive actions (not original) & Stay there. (2.8) Stay calmly. (2.0) \\ \hline
Negative actions (not original) & Screw it. I'm going carefully. (-17.4) Yell at everyone. (-13.5) \\ \hline
Irrelevant actions (not original) & Insert a coin. (-1.4) Throw a coin to the ground. (-3.6) \\ \hline
\end{tabular}
\caption{Predicted Q-value examples}
\label{table:q_value_examples}
\end{center}
\end{table*}

We use the baseline models and proposed DRRN model trained with the original action descriptions for ``Machine of Death'', and test on paraphrased action descriptions. For this game, the underlying state transition mechanism has not changed. The only change to the game interface is that during testing, every time the player reads the actions from the game simulator, it reads the paraphrased descriptions and performs selection based on these paraphrases. Since the texts in test time are ``unseen'' to the player, a good model needs to have some level of language understanding, while a naive model that memorizes all unique action texts in the original game will do poorly. The results for these models are shown in Table \ref{table:final-perf-game2-paraphrase}. All methods have a slightly lower average reward in this setting (10.5 vs. 11.2 for the original actions), but the DRRN still gives a high reward and significantly outperforms other methods. This shows that the DRRN can generalize well to ``unseen'' natural language descriptions of actions.

\begin{table}[t]
\begin{center}
\footnotesize
\begin{tabular}{| p{2.25cm} | >{\centering}p{1.20cm} | >{\centering}p{1.20cm} | c |} \hline
\rowcolor{Gray} Eval metric & \multicolumn{3}{c|}{Average reward} \\ \hline
hidden dimension & 20 & 50 & 100 \\ \hline
PA DQN ($L=2$) & 0.2 (1.2) & 2.6 (1.0) & 3.6 (0.3)  \\ \hline
MA DQN ($L\!=\!2$) & 2.5 (1.3) & 4.0 (0.9) & 5.1 (1.1) \\ \hline
DRRN ($L=2$) & 7.3 (0.7) & 8.3 (0.7) & {\bf 10.5 (0.9)} \\ \hline
\end{tabular}
\caption{The final average rewards and standard deviations on paraphrased game ``Machine of Death''.}
\label{table:final-perf-game2-paraphrase}
\end{center}
\end{table}

%% file: related.tex
There has been increasing interest in applying deep reinforcement learning to a variety problems, but only a few studies address problems with natural language state or action spaces. 
 In language processing, reinforcement learning has been applied to a dialogue management system that converses with a human user by taking actions that generate natural language \cite{scheffler2002automatic,young2013pomdp}. There has also been interest in extracting textual knowledge to improve game control performance \cite{branavan2011learning}, and mapping text instructions to sequences of executable actions \cite{branavan-EtAl:2009:ACLIJCNLP}. In some applications, it is possible to manually design features for state-action pairs, which are then used in reinforcement learning to learn a near-optimal policy \cite{li2009reinforcement}.  Designing such features, however, require substantial domain knowledge.

The work most closely related to our study inolves application of deep reinforcement to learning decision policies for parser-based text games. Narasimhan et al.~\shortcite{narasimhan-kulkarni-barzilay:2015:EMNLP} applied a Long Short-Term Memory DQN framework, which achieves higher average reward than the random and Bag-of-Words DQN baselines. In this work, actions are constrained to a set of known fixed command structures (one action and one argument object), based on a limited action-side vocabulary size. The overall action space is defined by the action-argument product space. This pre-specified product space is not feasible for the more complex text strings in other forms of text-based games. Our proposed DRRN, on the other hand, can handle the more complex text strings, as well as parser-based games. In preliminary experiments with the parser-based game from \cite{narasimhan-kulkarni-barzilay:2015:EMNLP}, we find that the DRRN using a bag-of-words (BOW) input achieves results on par with their BOW DQN. The main advantage of the DRRN is that it can also handle actions described with more complex language.


The DRRN experiments described here leverage only a simple bag-of-words representation of phrases and sentences.    As observed in \cite{narasimhan-kulkarni-barzilay:2015:EMNLP}, more complex sentence-based models can give further improvements. In preliminary experiments with ``Machine of Death'', we did not find LSTMs to give improved performance, but we conjecture that they would be useful in larger-scale tasks, or when the word embeddings are initialized by training on large data sets.

As mentioned earlier, other work has applied deep reinforcement learning to a problem with a continuous action space \cite{lillicrap2015continuous}. In the DRRN, the action space is inherently discrete, but we learn a continuous representation of it. As indicated by the paraphrasing experiment, the continuous space representation seems to generalize reasonably well.


%% file: appendices.tex
\appendix

\section{Percentage of Choice-based and Hypertext-based Text Games}
\label{Appendix:choice-hypertext-popularity}
As shown in Table \ref{table:html_perc}.\footnote{Statistics are obtained from http://www.ifarchive.org}
\begin{table}[h]
\begin{center}
\footnotesize
\begin{tabular}{| l || p{1cm} | p{1cm} | p{1cm} | p{1cm} | p{1cm} |} \hline
Year & 2010 & 2011 & 2012 & 2013 & 2014 \\ \hline
Percentage & 7.69\% & 7.89\% & 25.00\% & 55.56\% & 61.90\% \\ \hline
\end{tabular}
\caption{Percentage of choice-based and hypertext-based text games since 2010, in archive of interactive fictions}
\label{table:html_perc}
\end{center}
\end{table}

\section{Back Propagation Formula for Learning DRRN}
\label{Appendix:BackPropDRRN}
Let $h_{l,s}$ and $h_{l,a}$ denote the $l$-th hidden layer for state and action side neural networks, respectively. For state side, $W_{l,s}$ and $b_{l,s}$ denote the linear transformation weight matrix and bias vector between the $(l-1)$-th and $l$-th hidden layers. For actions side, $W_{l,a}$ and $b_{l,a}$ denote the linear transformation weight matrix and bias vector between the $(l-1)$-th and $l$-th hidden layers. The DRRN has $L$ hidden layers on each side.

{\bf Forward:}
	\begin{align}
		h_{1,s}			&=	f(W_{1,s} s_t + b_{1,s}) \\
		h_{1,a}^i			&=	f(W_{1,a} a_t^i + b_{1,a}), \quad i=1, 2, 3, ..., |\mathcal{A}_t| \\
		h_{l,s}			&=	f(W_{l-1,s} h_{l-1,s} + b_{l-1,s}), \quad l=2, 3, ..., L \\
		h_{l,a}^i			&=	f(W_{l-1,a} h_{l-1,a}^i + b_{l-1,a}), \quad i=1, 2, 3, ..., |\mathcal{A}_t|, l=2, 3, ..., L \\
		Q(s_t, a_t^i)	&=	h_{L,s}^T h_{L,a}^i
	\end{align}
where $f(\cdot)$ is the nonlinear activation function at the hidden layers, which is chosen as $\tanh{(x)}=(1-\exp{(-2x)})/(1+\exp{(-2x)})$, and $\mathcal{A}_t$ denotes the set of all actions at time $t$.

{\bf Backward:}

Note we only back propagate for actions that are actually taken. More formally, let $a_t$ be action the DRRN takes at time $t$, and denote $\Delta=[Q(s_t,a_t)-(r_t+\gamma \max_a{Q(s_{t+1},a)})]^2/2$. Denote $\delta_{l,s} = \delta b_{l,s} = \partial Q/\partial b_{l,s}$, $\delta_{l,a} = \delta b_{l,a} = \partial Q/\partial b_{l,a}$, and we have (by following chain rules):
\begin{align}
	\delta Q = \frac{\partial{\Delta}}{\partial{Q}}=Q(s_t,a_t)-(r_t+\gamma \max_a{Q(s_{t+1},a)})
\end{align}
\begin{align}
	\begin{cases}
		\delta_{L,s} = \delta Q \cdot h_{L,a} \odot (1-h_{L,s}) \odot (1+h_{L,s})$$ \\
		\delta_{l-1,s} = W_{l,s}^T \delta_{l,s} \odot (1-h_{l-1,s}) \odot (1+h_{l-1,s}), \quad l=2, 3, ..., L
	\end{cases}
\end{align}
\begin{align}
	\begin{cases}
		\delta_{L,a} = \delta Q \cdot h_{L,s} \odot (1-h_{L,a}) \odot (1+h_{L,a})$$ \\
		\delta_{l-1,a} = W_{l,a}^T \delta_{l,a} \odot (1-h_{l-1,a}) \odot (1+h_{l-1,a}), \quad l=2, 3, ..., L
	\end{cases}
\end{align}
\begin{align}
	\begin{cases}
		\delta W_{1,s} = \partial Q/\partial W_{1,s} = \delta_{1,s} \cdot s_t^T \\
		\delta W_{l,s} = \partial Q/\partial W_{l,s} = \delta_{l,s} \cdot h_{l-1,s}^T, \quad l=2, 3, ..., L
	\end{cases}
\end{align}
\begin{align}
	\begin{cases}
		\delta W_{1,a} = \partial Q/\partial W_{1,a} = \delta_{1,a} \cdot a_t^T \\
		\delta W_{l,a} = \partial Q/\partial W_{l,a} = \delta_{l,a} \cdot h_{l-1,a}^T, \quad l=2, 3, ..., L
	\end{cases}
\end{align}
where $\odot$ denotes element-wise Hadamard product.

\section{Final Rewards in the Two Text Games}
\label{Appendix:DefineRewards}
As shown in Table \ref{table:define-rewards} and Table \ref{table:define-rewards-game2}.
\begin{table}[t]
\begin{center}
\footnotesize
\begin{tabular}{| c | p{12cm} |} \hline
\rowcolor{Gray} Reward & Endings (partially shown) \\ \hline
-20 & Suspicion fills my heart and I scream. Is she trying to kill me? I don't trust her one bit... \\ \hline
-10 & Submerged under water once more, I lose all focus...  \\ \hline
0 & Even now, she's there for me. And I have done nothing for her... \\ \hline
10 & Honest to God, I don't know what I see in her. Looking around, the situation's not so bad... \\ \hline
20 & Suddenly I can see the sky... I focus on the most important thing - that I'm happy to be alive.\\ \hline
\end{tabular}
\caption{Final rewards defined for the text game ``Saving John"}
\label{table:define-rewards}
\end{center}
\end{table}

\begin{table}[t]
\begin{center}
\footnotesize
\begin{tabular}{| c | p{12cm} |} \hline
\rowcolor{Gray}  Reward & Endings (partially shown) \\ \hline
-20 & You spend your last few moments on Earth lying there, shot through the heart, by the image of Jon Bon Jovi. \\ \hline
-20 & you hear Bon Jovi say as the world fades around you.\\ \hline
-20 & As the screams you hear around you slowly fade and your vision begins to blur, you look at the words which ended your life. \\ \hline
-10 & You may be locked away for some time. \\ \hline
-10 & Eventually you're escorted into the back of a police car as Rachel looks on in horror. \\ \hline
-10 & Fate can wait. \\ \hline
-10 & Sadly, you're so distracted with looking up the number that you don't notice the large truck speeding down the street. \\ \hline
-10 & All these hiccups lead to one grand disaster. \\ \hline
10 & Stay the hell away from me! She blurts as she disappears into the crowd emerging from the bar. \\ \hline
20 & You can't help but smile. \\ \hline
20 & Hope you have a good life.\\ \hline
20 & Congratulations! \\ \hline
20 & Rachel waves goodbye as you begin the long drive home. After a few minutes, you turn the radio on to break the silence. \\ \hline
30 & After all, it's your life. It's now or never. You ain't gonna live forever. You just want to live while you're alive. \\ \hline
\end{tabular}
\caption{Final rewards for the text game ``Machine of Death.'' Scores are assigned according to whether the character survives, how the friendship develops, and whether he overcomes his fear.}
\label{table:define-rewards-game2}
\end{center}
\end{table}

\section{Game 2 Learning curve with shared state and action embedding}
\label{Appendix:learning-curve-shared}
\begin{figure*}[t]
\centerline{
		\includegraphics[width=0.48\textwidth]{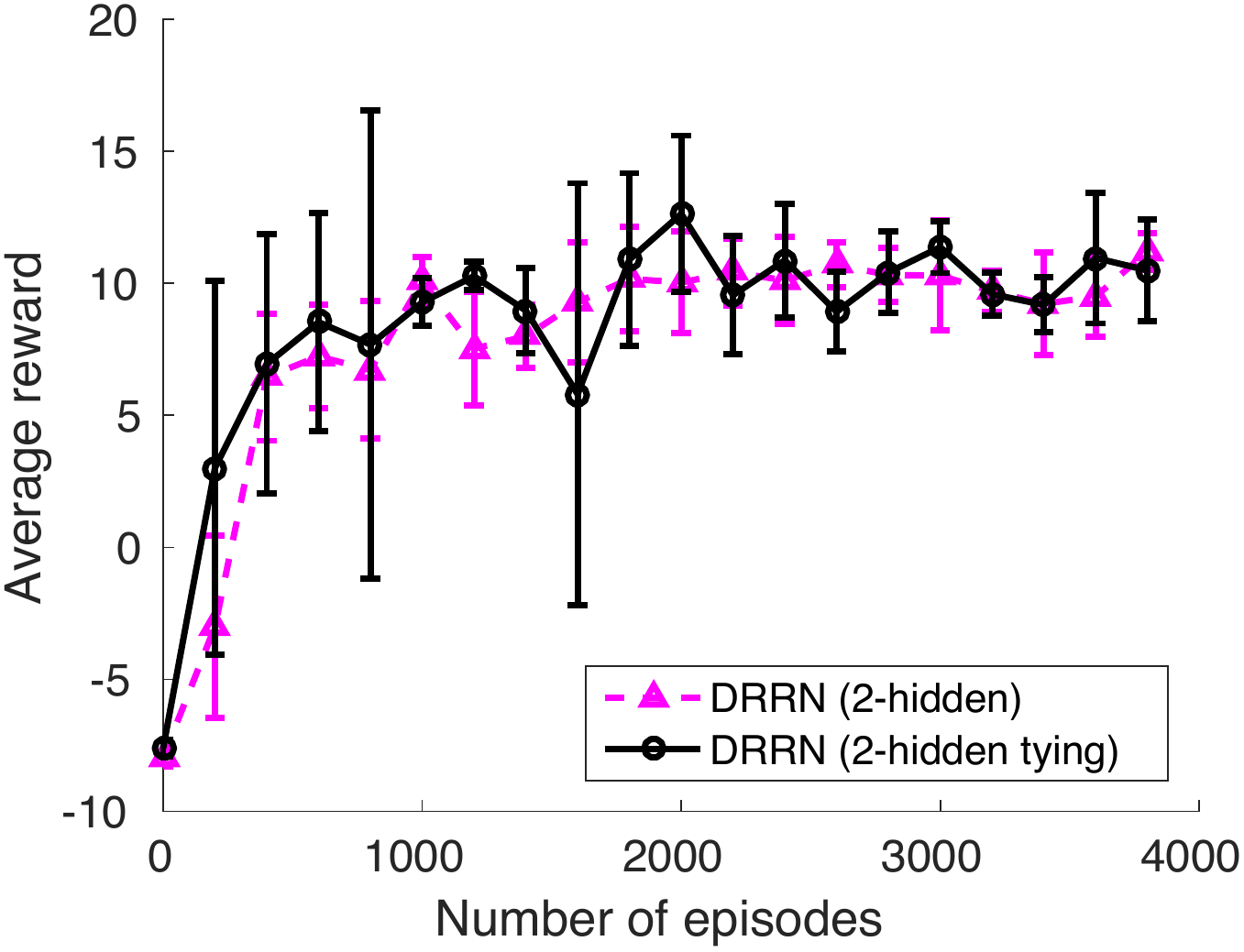}
 } 
  \caption{Learning curves of shared state-action embedding vs. proposed DRRN in Game 2}
  \label{Fig:LearningCurveShared}
\end{figure*}

As shown in Figure \ref{Fig:LearningCurveShared}. For the first 1000 episodes, parameter tying gives faster convergence, but learning curve also has high variance and unstable.

\section{Examples of State-Action Pairs in the Two Text Games}
\label{Appendix:text-pair-Qvalues}
As shown in Table \ref{table:game1-visual} and Table \ref{table:game2-visual}.
\begin{table}[t]
\begin{center}
\footnotesize
\begin{tabular}{| p{8.4cm} | p{4.4cm} |} \hline
\rowcolor{Gray} State & Actions (with Q values) \\ \hline
A wet strand of hair hinders my vision and I'm back in the water. Sharp pain pierces my lungs. How much longer do I have? 30 seconds? Less? I need to focus. A hand comes into view once more. & I still don't know what to do. (-8.981) Reach for it. (18.005) \\ \hline
''Me:'' Hello Sent: today ''Cherie:'' Hey. Can I call you? Sent: today & Reply "I'll call you" (14.569) No (-9.498) \\ \hline
"You don't hold any power over me. Not anymore." Lucretia raises one eyebrow. The bar is quiet. "I really wish I did my hair today." She twirls a strand. "I'm sorry," "Save it." //Yellow Submarine plays softly in the background.// "I really hate her." "Cherie? It's not her fault." "You'll be sorry," "Please stop screaming." & I laugh and she throws a glass of water in my face. (16.214) I look away and she sips her glass quietly. (-7.986) \\ \hline
My dad left before I could remember. My mom worked all the time but she had to take care of her father, my grandpa. The routine was that she had an hour between her morning shift and afternoon shift, where she'd make food for me to bring to pops. He lived three blocks away, in a house with red steps leading up to the metal front door. Inside, the stained yellow wallpaper and rotten oranges reeked of mold. I'd walk by myself to my grandfather's and back. It was lonely sometimes, being a kid and all, but it was nothing I couldn't deal with. It's not like he abused me, I mean it hurt but why wouldn't I fight back? I met Adam on one of these walks. He made me feel stronger, like I can face anything. & Repress this memory (-8.102) Why didn't I fight back? (10.601) Face Cherie (14.583) \\ \hline
\end{tabular}
\caption{Q values (in parentheses) for state-action pair from ``Saving John'', using trained DRRN. High Q-value actions are more cooperative actions thus more likely leading to better endings}
\label{table:game1-visual}
\end{center}
\end{table}

\begin{table}[t]
\begin{center}
\footnotesize
\begin{tabular}{| p{8.4cm} | p{4.4cm} |} \hline
\rowcolor{Gray} State & Actions (with Q values) \\ \hline
Peak hour ended an hour or so ago, alleviating the feeling of being a tinned sardine that?s commonly associated with shopping malls, though there are still quite a few people busily bumbling about.  To your left is a fast food restaurant. To the right is a UFO catcher, and a poster is hanging on the wall beside it. Behind you is the one of the mall's exits.  In front of you stands the Machine.  You're carrying 4 dollars in change. & fast food restaurant (1.094) the Machine (3.708) mall's exits (0.900) UFO catcher (2.646) poster (1.062) \\ \hline
You lift the warm mug to your lips and take a small sip of hot tea. & Ask what he was looking for. (3.709) Ask about the blood stains. (7.488) Drink tea. (5.526) Wait. (6.557) \\ \hline
As you move forward, the people surrounding you suddenly look up with terror in their faces, and flee the street. & Ignore the alarm of others and continue moving forward. (-21.464) Look up. (16.593) \\ \hline
Are you happy? Is this what you want to do? If you didn't avoid that sign, would you be satisfied with how your life had turned out?  Sure, you're good at your job and it pays well, but is that all you want from work?   If not, maybe it's time for a change. & Screw it. I'm going to find a new life right now. It's not going to be easy, but it's what I want. (23.205) Maybe one day. But I'm satisfied right now, and I have bills to pay. Keep on going. (One minute) (14.491) \\ \hline
You slam your entire weight against the man, making him stumble backwards and drop the chair to the ground as a group of patrons race to restrain him.  You feel someone grab your arm, and look over to see that it?s Rachel.  Let's get out of here,  she says while motioning towards the exit.  You charge out of the bar and leap back into your car, adrenaline still pumping through your veins. As you slam the door, the glove box pops open and reveals your gun. & Grab it and hide it in your jacket before Rachel can see it. (21.885) Leave it. (1.915) \\ \hline
\end{tabular}
\caption{Q values (in parentheses) for state-action pair from ``Machine of Death'', using trained DRRN}
\label{table:game2-visual}
\end{center}
\end{table}

\section{Examples of State-Action Pairs that do not exist in the feasible set}
\label{Appendix:made-up-text-pair-Qvalues}
As shown in Table \ref{table:game2-visual-nonexisting-actions}.
\begin{table}[t]
\begin{center}
\footnotesize
\begin{tabular}{| p{6cm} | p{8cm} |} \hline
\rowcolor{Gray} & Text (with Q-values) \\ \hline
State & As you move forward, the people surrounding you suddenly look up with terror in their faces, and flee the street. \\ \hline
Actions that are in the feasible set & Ignore the alarm of others and continue moving forward. (-21.5) Look up. (16.6) \\ \hline
Positive actions that are not in the feasible set & Stay there. (2.8) Stay calmly. (2.0) \\ \hline
Negative actions that are not in the feasible set & Screw it. I'm going carefully. (-17.4) Yell at everyone. (-13.5) \\ \hline
Irrelevant actions that are not in the feasible set & Insert a coin. (-1.4) Throw a coin to the ground. (-3.6) \\ \hline
\end{tabular}
\caption{Q values (in parentheses) for state-action pair from ``Machine of Death'', using trained DRRN, with made-up actions that were not in the feasible set}
\label{table:game2-visual-nonexisting-actions}
\end{center}
\end{table}